%% file: CLGP-arXiv.tex
\documentclass{article}

\usepackage{times}
\usepackage{graphicx} 
\usepackage[caption=false]{subfig}

\usepackage{natbib}

\usepackage{algorithm}
\usepackage{algorithmic}

\usepackage[accepted]{icml2015arxiv}

\usepackage{url}
\usepackage{amsmath,amssymb}
\allowdisplaybreaks
\usepackage{algorithm}
\usepackage{graphicx}
\usepackage{color}
\usepackage{xspace}
\usepackage{array}
\usepackage{xcolor}
\usepackage{defs}
\usepackage{tabularx}

\newcommand\gp{\mathrm{GP}\xspace}

\DeclareMathOperator*{\softmax}{Softmax}
\DeclareMathOperator*{\categorical}{Categorical}
\DeclareMathOperator*{\lse}{lse}
\DeclareMathOperator*{\kl}{KL}

\usepackage[small,compact]{titlesec}
\titlespacing\section{0pt}{0pt plus 2pt minus 2pt}{0pt plus 0pt minus 0pt}
\titlespacing\subsection{0pt}{0pt plus 2pt minus 2pt}{0pt plus 0pt minus 0pt}
\titlespacing\subsubsection{0pt}{0pt plus 2pt minus 2pt}{0pt plus 0pt minus 0pt}
\expandafter\def\expandafter\normalsize\expandafter{%
    \normalsize
    \setlength\abovedisplayskip{0pt}
    \setlength\belowdisplayskip{0pt}
    \setlength\abovedisplayshortskip{0pt}
    \setlength\belowdisplayshortskip{0pt}
}

\usepackage{listings}
\usepackage{color}
\usepackage{setspace}

\usepackage{hyperref}

\definecolor{Code}{rgb}{0,0,0}
\definecolor{Decorators}{rgb}{0.5,0.5,0.5}
\definecolor{Numbers}{rgb}{0.5,0,0}
\definecolor{MatchingBrackets}{rgb}{0.25,0.5,0.5}
\definecolor{Keywords}{rgb}{0,0,1}
\definecolor{self}{rgb}{0,0,0}
\definecolor{Strings}{rgb}{0,0.63,0}
\definecolor{Comments}{rgb}{0,0.63,1}
\definecolor{Backquotes}{rgb}{0,0,0}
\definecolor{Classname}{rgb}{0,0,0}
\definecolor{FunctionName}{rgb}{0,0,0}
\definecolor{Operators}{rgb}{0,0,0}
\definecolor{Background}{rgb}{0.98,0.98,0.98}

\lstnewenvironment{python}[1][]{
\lstset{
numbers=left,
numberstyle=\footnotesize,
numbersep=1em,
xleftmargin=1em,
framextopmargin=2em,
framexbottommargin=2em,
showspaces=false,
showtabs=false,
showstringspaces=false,
frame=l,
tabsize=4,
basicstyle=\ttfamily\small,
backgroundcolor=\color{Background},
language=Python,
commentstyle=\color{Comments}\slshape,
stringstyle=\color{Strings},
morecomment=[s][\color{Strings}]{"""}{"""},
morecomment=[s][\color{Strings}]{'''}{'''},
morekeywords={import,from,class,def,for,while,if,is,in,elif,else,not,and,or,print,break,continue,return,True,False,None,access,as,,del,except,exec,finally,global,import,lambda,pass,print,raise,try,assert},
keywordstyle={\color{Keywords}\bfseries},
morekeywords={[2]@invariant},
keywordstyle={[2]\color{Decorators}\slshape},
emph={self},
emphstyle={\color{self}\slshape},
}}{}

\icmltitlerunning{Latent Gaussian Processes for Distribution Estimation of Multivariate Categorical Data}

\begin{document} 

\twocolumn[
\icmltitle{Latent Gaussian Processes for Distribution Estimation of Multivariate Categorical Data}

\vspace{-4mm}
\icmlauthor{Yarin Gal}{yg279@cam.ac.uk}
\icmlauthor{Yutian Chen}{yc373@cam.ac.uk}
\icmlauthor{Zoubin Ghahramani}{zoubin@cam.ac.uk}
\icmladdress{University of Cambridge}

\vspace{-8mm}

\icmlkeywords{boring formatting information, machine learning, ICML}

\vskip 0.3in
]

\input{CLGP-paper}

\include{CLGP-appendix}
\end{document}

%% file: CLGP-paper.tex
\begin{abstract} 
Multivariate categorical data occur in many applications of machine learning. One of the main difficulties with these vectors of categorical variables is sparsity. The number of possible observations grows exponentially with vector length, but dataset diversity might be poor in comparison.
Recent models have gained significant improvement in \textit{supervised} tasks with this data. 
These models embed observations in a continuous space to capture similarities between them. 
Building on these ideas we propose a Bayesian model for the \textit{unsupervised} task of distribution estimation of multivariate categorical data. 

We model vectors of categorical variables as generated from a non-linear transformation of a continuous latent space. Non-linearity captures multi-modality in the distribution. The continuous representation addresses sparsity. 
Our model ties together many existing models, linking the linear categorical latent Gaussian model, the Gaussian process latent variable model, and Gaussian process classification. 
We derive inference for our model based on recent developments in sampling based variational inference. We show empirically that the model outperforms its linear and discrete counterparts in imputation tasks of sparse data.
\end{abstract}

\vspace{-4mm}
\section{Introduction}

Multivariate categorical data is common in fields ranging from language processing to medical diagnosis. Recently proposed supervised models have gained significant improvement in tasks involving \textit{big labelled data} of this form (see for example \citet{bengio2006neural,collobert2008unified}). These models rely on information that had been largely ignored before: similarity between vectors of categorical variables. But what should we do in the unsupervised setting, when we face \textit{small unlabelled data} of this form?

Medical diagnosis provides good examples of small unlabelled data. Consider a dataset composed of test results of a relatively small number of patients. Each patient has a medical record composed often of dozens of examinations, taking various categorical test results. We might be faced with the task of deciding which tests are necessary for a patient under examination to take, and which examination results could be deduced from the existing tests. This can be achieved with \textit{distribution estimation}.

Several tools in the Bayesian framework could be used for this task of distribution estimation of unlabelled small datasets. Tools such as the Dirichlet-Multinomial distribution and its extensions are an example of such. These rely on relative frequencies of categorical variables appearing with others, with the addition of various smoothing techniques. But when faced with long multivariate sequences, these models run into problems of sparsity. This occurs when the data consists of vectors of categorical variables with most configurations of categorical values not in the dataset. In medical diagnosis this happens when there is a large number of possible examinations compared to a small number of patients.

Building on ideas used for big labelled discrete data, we propose a Bayesian model for distribution estimation of small unlabelled data. Existing supervised models for discrete labelled data embed the observations in a continuous space. This is used to find the similarity between vectors of categorical variables. We extend this idea to the small unlabelled domain by modelling the continuous embedding as a latent variable. A generative model is used to find a distribution over the discrete observations by modelling them as dependent on the continuous latent variables.

Following the medical diagnosis example, patient $n$ would be modelled by a continuous latent variable $x_n \in \cX$. For each examination $d$, the latent $x_n$ induces a vector of probabilities $\bff = (f_{nd1}, ..., f_{ndK})$, one probability for each possible test result $k$. A categorical distribution returns test result $y_{nd}$ based on these probabilities, resulting in a patient's medical assessment $y_n = (y_{n1}, ..., y_{nD})$. We need to decide how to model the distribution over the latent space $\cX$ and vectors of probabilities $\bff$.


We would like to capture sparse multi-modal categorical distributions. A possible approach would be to model the continuous representation with a simple latent space and a non-linear transformation of points in the space to obtain probabilities. In this approach we place a standard normal distribution prior on a latent space, and feed the output of a non-linear transformation of the latent space into a Softmax to obtain probabilities. 
We use sparse Gaussian processes (GPs) to transform the latent space non-linearly. Sparse GPs form a distribution over functions supported on a small number of points with linear time complexity \citep{quinonero2005unifying,Titsias09variationallearning}. 
We use a covariance function that is able to transform the latent space non-linearly. We name this model the \textit{Categorical Latent Gaussian Process} (CLGP).
Using a Gaussian process with a linear covariance function recovers the linear Gaussian model \citep[LGM, ][]{marlin2011piecewise}. This model linearly transform a continuous latent space resulting in discrete observations. 

The Softmax likelihood is not conjugate to the our Gaussian prior, and integrating the latent variables with a Softmax distribution is intractable. 
A similar problem exists with LGMs. \citet{marlin2011piecewise} solved this by using variational inference and various bounds for the likelihood in the binary case, or alternative likelihoods to the Softmax in the categorical case \citep{khan2012stick}. Many bounds have been studied in the literature for the binary case: Jaakkola and Jordan's bound \citep{Jaakkola96avariational}, the tilted bound \citep{knowles2011non}, piecewise linear and quadratic bounds \citep{marlin2011piecewise}, and others. But for categorical data fewer bounds exist, since the multivariate Softmax is hard to approximate in high-dimensions. The Bohning bound \citep{bohning1992multinomial} and Blei and Lafferty's bound \citep{blei2006correlated} give poor approximation \citep{khan2012stick}.

Instead we use recent developments in sampling-based variational inference \citep{blei2012variational} to avoid integrating the latent variables with the Softmax analytically.
Our approach takes advantage of this tool to obtain simple yet powerful model and inference. We use Monte Carlo integration to approximate the non-conjugate likelihood obtaining noisy gradients \citep{kingma2013auto,rezende2014stochastic,titsias2014doubly}. We then use learning-rate free stochastic optimisation \citep{Tieleman2012COURSERA} to optimise the noisy objective. We leverage symbolic differentiation \citep[Theano,][]{bergstra+al:2010-scipy} to obtain simple and modular code\footnote{Python code for the model and inference is given in the appendix, and available at \url{http://github.com/yaringal/CLGP}}. 

We experimentally show the advantages of using non-linear transformations for the latent space. We follow the ideas brought in \citet{paccanaro2001learning} and evaluate the models on relation embedding and relation learning. 
We then demonstrate the utility of the model in the real-world \textit{sparse data} domain. We use a medical diagnosis dataset where data is scarce, comparing our model to discrete frequency based models. 
We use the estimated distribution for a task similar to the above, where we attempt to infer which test results can be deduced from the others. 
We compare the models on the task of imputation of raw data studying the effects of government terror warnings on political attitudes. We then evaluate the continuous models on a binary Alphadigits dataset composed of binary images of handwritten digits and letters, where each class contains a small number of images. We inspect the latent space embeddings and separation of the classes.
Lastly, we evaluate the robustness of our inference, inspecting the Monte Carlo estimate variance over time. 

\section{Related Work}

\begin{figure}[b!]
\hspace{-4mm}
  \includegraphics[trim = 18mm 30mm 5mm 25mm, clip, width=1.10\linewidth]{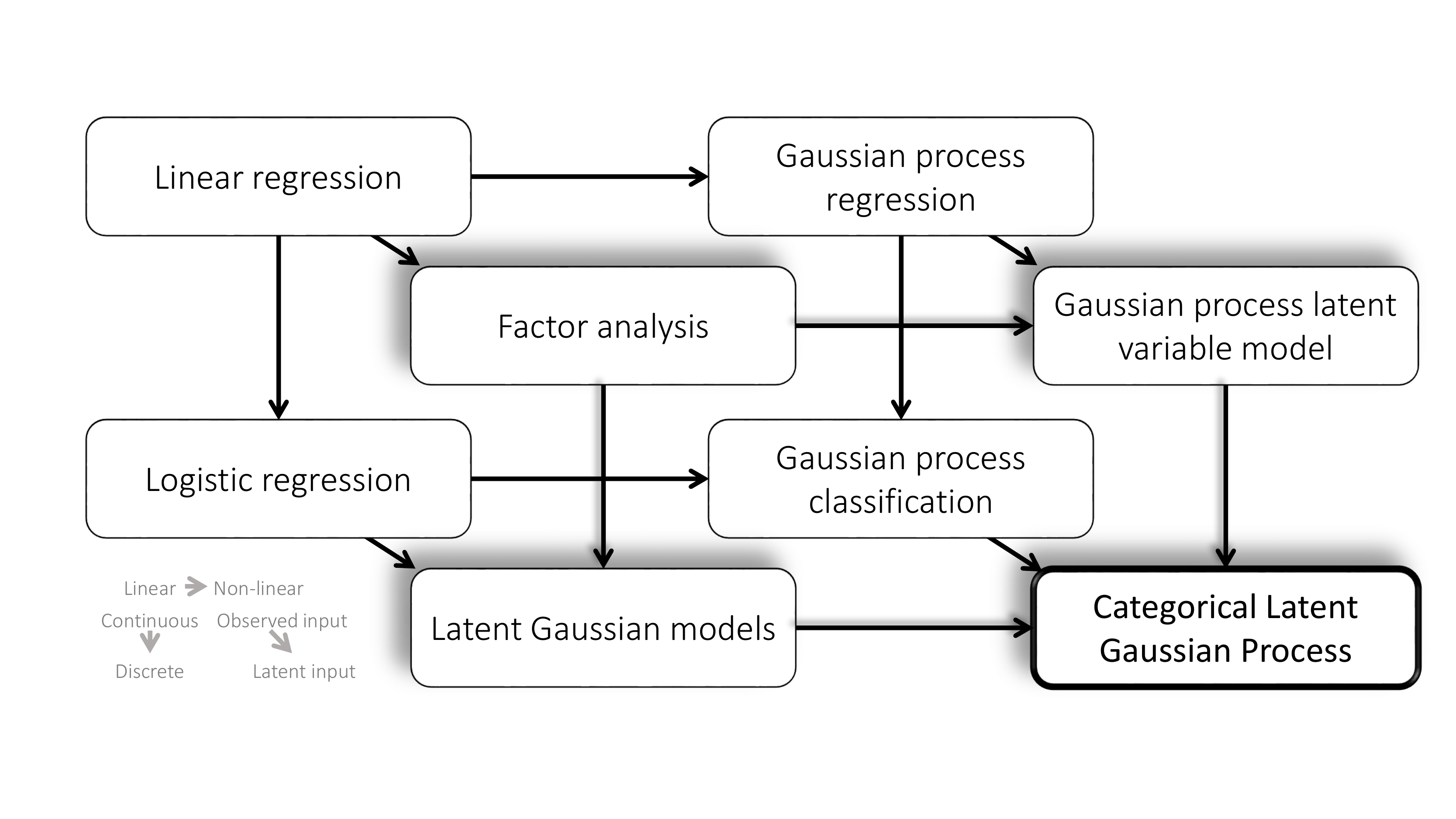}
  \caption{Relations between existing models and the model proposed in this paper (\textit{Categorical Latent Gaussian Process}); the model can be seen as a non-linear version of the \textit{latent Gaussian model} (left to right, \citet{khan2012stick}), it can be seen as a latent counterpart to the \textit{Gaussian process classification} model (back to front, \citet{rasmussen2006gaussian}), or alternatively as a discrete extension of the \textit{Gaussian process latent variable model} (top to bottom, \citet{lawrence2005probabilistic}).} \label{fig:relations}
\end{figure}

Our model (CLGP) relates to some key probabilistic models (fig.\ \ref{fig:relations}). It can be seen as a non-linear version of the \textit{latent Gaussian model} (LGM, \citet{khan2012stick}) as discussed above. In the LGM we have a standard normal prior placed on a latent space, which is transformed linearly and fed into a Softmax likelihood function. The probability vector output is then used to sample a single categorical value for each categorical variable (e.g.\ medical test results) in a list of categorical variables (e.g.\ medical assessment). These categorical variables correspond to elements in a multivariate categorical vector. The parameters of the linear transformation are optimised directly within an EM framework. \citet{khan2012stick} avoid the hard task of approximating the Softmax likelihood by using an alternative function (product of sigmoids) which is approximated using numerical techniques. Our approach avoids this cumbersome inference. 

Our proposed model can also be seen as a latent counterpart to the \textit{Gaussian process classification} model \citep{rasmussen2006gaussian}, in which a Softmax function is again used to discretise the continuous values. The continuous valued outputs are obtained from a Gaussian process, which non-linearly transforms the inputs to the classification problem. Compared to GP classification where the inputs are fully observed, in our case the inputs are latent.
Lastly, our model can be seen as a discrete extension of the \textit{Gaussian process latent variable model} \citep[GPLVM,][]{lawrence2005probabilistic}. This model has been proposed recently as means of performing non-linear dimensionality reduction (counterpart to the linear principal component analysis \citep{tipping1999probabilistic}) and density estimation in continuous space. 

\section{A Latent Gaussian Process Model for Multivariate Categorical Data}
We consider a generative model for a dataset $\bY$ with $N$ observations (patients for example) and $D$ categorical variables (different possible examinations). The $d$-th categorical variable in the $n$-th observation, $y_{nd}$, is a categorical variable that can take an integer value from $0$ to $K_d$. For ease of notation, we assume all the categorical variables have the same cardinality, i.e.\ $K_d\equiv K,~\forall d=1,\dots,D$. 

In our generative model, each categorical variable $y_{nd}$ follows a categorical distribution with probability given by a Softmax with weights $\bff_{nd}=(0,f_{nd1},\dots,f_{ndK})$. Each weight $f_{ndk}$ is the output of a nonlinear function of a $Q$ dimensional latent variable $\bx_{n} \in \eR^Q$: $\cF_{dk}(\bx_{n})$. To complete the generative model, we assign an isotropic Gaussian distribution prior with standard deviation $\sigma_x^2$ for the latent variables, and a Gaussian process prior for each of the nonlinear functions. We also consider a set of $M$ auxiliary variables which are often called inducing inputs. These inputs $\bZ \in \eR^{M \times Q}$ lie in the latent space with their corresponding outputs $\bU \in \eR^{M \times D \times K}$ lying in the weight space (together with $f_{ndk}$). The inducing points are used as ``support'' for the function. Evaluating the covariance function of the GP on these instead of the entire dataset allows us to perform approximate inference in $\mathcal{O}(M^2 N)$ time complexity instead of $\mathcal{O}(N^3)$ (where $M$ is the number of inducing points and $N$ is the number of data points \citep{quinonero2005unifying}).

The model is expressed as:
\begin{align}
x_{nq} &\iiddist \cN(0,\sg_x^2)\\
\cF_{dk} &\iiddist \gp(\bK_d) \nn \\
f_{ndk} &= \cF_{dk}(\bx_n),\quad u_{mdk} = \cF_{dk}(\bz_m) \nn\\
y_{nd} &\sim \softmax(\bff_{nd}), \nn
\end{align}
for $n \in [N]$ (the set of naturals between 1 and $N$), $q \in [Q]$, $d \in [D]$, $k \in [K]$, $m \in [M]$, and
where the Softmax distribution is defined as,
\begin{align}
\softmax(y=k;\bff) &= \categorical \bigg( \frac{\exp(f_k)}{\exp(\lse(\bff))} \bigg), \nn\\
\lse(\bff) &= \log\bigg(1+\sum_{k'=1}^K \exp(f_{k'})\bigg), \label{eqn:lse}
\end{align}
for $k=0,\dots,K$ and with $f_0 :=0$.

Following our medical diagnosis example from the introduction, each patient is modelled by latent $\bx_n$; for each examination $d$, $\bx_n$ has a sequence of weights $(f_{nd1}, ..., f_{ndK})$, one weight for each possible test result $k$, that follows a Gaussian process; Softmax returns test result $y_{nd}$ based on these weights, resulting in a patient's medical assessment $\by_{n} = (y_{n1}, ..., y_{nD})$.


Define $\bff_{dk} = (f_{1dk}, ..., f_{Ndk})$ and define $\bu_{dk} = (u_{1dk}, ..., u_{Mdk})$.
The joint distribution of $(\bff_{dk}, \bu_{dk})$ with the latent nonlinear function, $\cF_{dk}$, marginalized under the GP assumption, is a multi-variate Gaussian distribution $\cN(\bzero, \bK_d([\bX, \bZ], [\bX, \bZ]))$. It is easy to verify that when we further marginalize the inducing outputs, we end up with a joint distribution of the form $\bff_{dk} \sim \cN(\bzero, \bK_d(\bX, \bX)), \forall d, k$. Therefore, the introduction of inducing outputs does not change the marginal likelihood of the data $\bY$. These are used in the variational inference method in the next section and the inducing inputs $\bZ$ are considered as variational parameters. 

We use the automatic relevance determination (ARD) radial basis function (RBF) covariance function 
for our model.
ARD RBF is able to select the dimensionality of the latent space automatically and transform it non-linearly.

\section{Inference}

The marginal log-likelihood, also known as log-evidence, is intractable for our model due to the non-linearity of the covariance function of the GP and the Softmax likelihood function. We first describe a lower bound of the log-evidence (ELBO) by applying Jensen's inequality with a variational distribution of the latent variables following \citet{titsias2010bayesian}.

Consider a variational approximation to the posterior distribution of $\bX$, $\bF$ and $\bU$ factorized as follows:
\begin{equation}
q(\bX,\bF,\bU) = q(\bX) q(\bU) p(\bF|\bX,\bU).
\end{equation}

We can obtain the ELBO by applying Jensen's inequality
\begin{align}
\log p(\bY) &= \log \int p(\bX) p(\bU) p(\bF|\bX,\bU) p(\bY | \bF) \d\bX \d\bF \d\bU \nn\\
&\geq \int q(\bX) q(\bU) p(\bF|\bX,\bU) \nn\\
&\qquad \cdot \log \frac{p(\bX) p(\bU) p(\bF|\bX,\bU) p(\bY | \bF)}{q(\bX) q(\bU) p(\bF|\bX,\bU)} \d\bX \d\bF \d\bU \nn\\
&= -\kl(q(\bX)\|p(\bX)) - \kl(q(\bU)\|p(\bU)) \nn\\
&\quad \quad + \sum_{n=1}^N \sum_{d=1}^D \int q(\bx_n) q(\bU_d) p(\bff_{nd}|\bx_n,\bU_d) \nn\\
&\qquad\qquad\qquad\qquad \cdot \log p(\by_{nd} | \bff_{nd}) \d\bx_n \d\bff_{nd} \d\bU_d \nn\\
&:= \cL \label{eqn:elbo}
\end{align}
where 
\begin{equation}
p(\bff_{nd}|\bx_n,\bU_d) = \prod_{k=1}^K \cN(f_{ndk} | \baa_{nd}^T \bu_{dk}, b_{nd}) \label{eq:f_cond}
\end{equation}
with
\begin{align}
\baa_{nd} &= \bK_{d,MM}^{-1} \bK_{d,Mn}, \nn\\
b_{nd} &= K_{d,nn} - \bK_{d,nM} \bK_{d,MM}^{-1} \bK_{d,Mn}. \label{eqn:a_b}
\end{align}

Notice however that the integration of $\log p(\by_{nd}|\bff_{nd})$ in eq. \ref{eqn:elbo} involves a nonlinear function ($\lse(\bff)$ from eq. \ref{eqn:lse}) and is still intractable. Consequently, we do not have an analytical form for the optimal variational distribution of $q(\bU)$ unlike in \citet{titsias2010bayesian}. Instead of applying a further approximation/lower bound on $\cL$, we want to obtain better accuracy by following a sampling-based approach \citep{blei2012variational,kingma2013auto,rezende2014stochastic,titsias2014doubly} to compute the lower bound $\cL$ and its derivatives with the Monte Carlo method. 

Specifically, we draw samples of $\bx_n$, $\bU_d$ and $\bff_{nd}$ from $q(\bx_n)$, $q(\bU_d)$, and $p(\bff_{nd}|\bx_n,\bU_d)$ respectively and estimate $\cL$ with the sample average. Another advantage of using the Monte Carlo method is that we are not constrained to a limited choice of covariance functions for the GP that is otherwise required for an analytical solution in standard approaches to GPLVM for continuous data \citep{titsias2010bayesian,hensman2013gaussian}.

We consider a mean field approximation for the latent points $q(\bX)$ as in \cite{titsias2010bayesian} and a joint Gaussian distribution with the following factorisation for $q(\bU)$:
\begin{align}
q(\bU) &= \prod_{d=1}^D \prod_{k=1}^K \cN(\bu_{dk} | \bmu_{dk}, \bSig_d), \nn\\
q(\bX) &= \prod_{n=1}^N \prod_{i=1}^Q \cN(x_{ni} | m_{ni}, s_{ni}^2)
\end{align}
where the covariance matrix $\bSig_d$ is shared for the same categorical variable $d$ (remember that $K$ is the number of values this categorical variable can take). The KL divergence in $\cL$ can be computed analytically with the given variational distributions. The parameters we need to optimise over\footnote{Note that the number of parameters to optimise over can be reduced by transforming the latent space non-linearly to a second low-dimensional latent space, which is then transformed linearly to the weight space containing points $f_{ndk}$.} include the hyper-parameters for the GP $\bta_d$, variational parameters for the inducing points $\bZ$, $\bmu_{dk}$, $\bSig_d$, and the mean and standard deviation of the latent points $m_{ni}$, $s_{ni}$.

\subsection{Transforming the Random Variables}
In order to obtain a Monte Carlo estimate to the gradients of $\cL$ with low variance, a useful trick introduced in \cite{kingma2013auto} is to transform the random variables to be sampled so that the randomness does not depend on the parameters with which the gradients will be computed. We present the transformation of each variable to be sampled as follows:

\paragraph{Transforming $\bX$.}
For the mean field approximation, the transformation for $\bX$ is straightforward as
\begin{equation}
x_{ni} = m_{ni} + s_{ni} \eps_{ni}^{(x)}, \quad
\eps_{ni}^{(x)} \sim \cN(0,1) \label{eq:eps_x}
\end{equation}

\paragraph{Transforming $\bu_{dk}$.}
The variational distribution of $\bu_{dk}$ is a joint Gaussian distribution. Denote the Cholesky decomposition of $\bSig_d$ as $\bL_d \bL_d^T = \bSig_d$. We can rewrite $\bu_{dk}$ as
\begin{equation}
\bu_{dk} = \bmu_{dk} + \bL_d \beps_{dk}^{(u)},\quad
\beps_{dk}^{(u)} \sim \cN(\bzero, \bI_M) \label{eq:eps_u}
\end{equation}
We optimize the lower triangular matrix $\bL_d$ instead of $\bSig_d$.

\paragraph{Transforming $\bff_{nd}$.}
Since the conditional distribution $p(\bff_{nd}|\bx_n,\bU_d)$ in Eq.~\ref{eq:f_cond} is factorized over $k$ we can define a new random variable for every $f_{ndk}$:
\begin{equation}
f_{ndk} = \baa_{nd}^T \bu_{dk} + \sqrt{b_{nd}} \eps_{ndk}^{(f)}, \quad
\eps_{ndk}^{(f)} \sim \cN(0,1) \label{eq:eps_f}
\end{equation}

Notice that the transformation of the variables does not change the distribution of the original variables and therefore does not change the KL divergence in Eq.~\ref{eq:f_cond}.

\subsection{Lower Bound with Transformed Variables}
Given the transformation we just defined, we can represent the lower bound as
\begin{align}
\cL =& -\sum_{n=1}^N \sum_{i=1}^Q \kl(q(x_{ni})\|p(x_{ni}))\nn\\
&\quad - \sum_{d=1}^D \sum_{k=1}^K \kl(q(\bu_{dk})\|p(\bu_{dk})) \nn\\
&\quad + \sum_{n=1}^N \sum_{d=1}^D \eE_{\beps_n^{(x)}, \beps_d^{(u)}, \beps_{nd}^{(f)}} \bigg[ \nn\\
&\quad \log \softmax \left(\by_{nd} \middle| \bff_{nd}\left(\beps_{nd}^{(f)}, \bU_{d}(\beps_d^{(u)}), \bx_{n}(\beps_n^{(x)})\right)\right) \bigg] \label{eqn:full_L_mf}
\end{align}
where the expectation in the last line is with respect to the fixed distribution defined in Eqs.~\ref{eq:eps_x}, \ref{eq:eps_u} and \ref{eq:eps_f}. Each expectation term that involves the Softmax likelihood, denoted as $\cL^{nd}_{s}$, can be estimated using Monte Carlo integration as
\begin{align}
&\cL^{nd}_s \approx \nn\\
&\quad \frac{1}{T} \sum_{i=1}^T
  \log \softmax \left(\by_{nd} \middle| \bff_{nd}\left(\beps_{nd,i}^{(f)}, \bU_{d}(\beps_{d,i}^{(u)}), \bx_{n}(\beps_{n,i}^{(x)})\right)\right)
\end{align}
with $\beps_{n,i}^{(x)}, \beps_{d,i}^{(u)}, \beps_{nd,i}^{(f)}$ drawn from their corresponding distributions. Since these distributions do not depend on the parameters to be optimized, the derivatives of the objective function $\cL$ are now straight-forward to compute with the same set of samples using the chain rule.

\subsection{Stochastic Gradient Descent}

We use gradient descent to find an optimal variational distribution. Gradient descent with noisy gradients is guaranteed to converge to a local optimum given decreasing learning rate with some conditions, but is hard to work with in practice. Initial values set for the learning rate influence the rate at which the algorithm converges, and misspecified values can cause it to diverge. For this reason new techniques have been proposed that handle noisy gradients well. Optimisers such as AdaGrad \citep{duchi2011adaptive}, AdaDelta \citep{zeiler2012adadelta}, and RMSPROP \citep{Tieleman2012COURSERA} have been proposed, each handling the gradients slightly differently, all averaging over past gradients. \citet{SchaulAS13} have studied empirically the different techniques, comparing them to one another on a variety of unit tests. They found that RMSPROP works better on many test sets compared to other optimisers evaluated. We thus chose to use RMSPROP for our experiments.

A major advantage of our inference is that it is extremely easy to implement and adapt. The straight-forward computation of derivatives through the expectation makes it possible to use symbolic differentiation. We use Theano \citep{bergstra+al:2010-scipy} for the inference implementation, where the generative model is implemented as in Eqs.~\ref{eq:eps_x}, \ref{eq:eps_u} and \ref{eq:eps_f}, and the optimisation objective, evaluated on samples from the generative model, is given by Eq. \ref{eqn:full_L_mf}.

\section{Experimental Results}

We evaluate the categorical latent Gaussian process (CLGP), comparing it to existing models for multivariate categorical distribution estimation. These include models based on a discrete latent representation (such as the Dirichlet-Multinomial), and continuous latent representation with a linear transformation of the latent space (latent Gaussian model \citep[LGM, ][]{khan2012stick}). We demonstrate over-fitting problems with the LGM, and inspect the robustness of our inference. 

For the following experiments, both the linear and non-linear models were initialised with a 2D latent space. The mean values of the latent points, $m_{n}$, were initialised at random following a standard normal distribution. The mean values of the inducing outputs ($\bmu_{dk}$) were initialised with a normal distribution with standard deviation $10^{-2}$. This is equivalent to using a uniform initial distribution for all values.
We initialise the standard deviation of each latent point ($s_{n}$) to 0.1, and initialise the length-scales of the ARD RBF covariance function to 0.1. We then optimise the variational distribution for 500 iterations. 
At every iteration we optimise the various quantities while holding $\bu_{dk}$'s variational parameters fixed, and then optimise $\bu_{dk}$'s variational parameters holding the other quantities fixed. 

Our setting supports semi-supervised learning with partially observed data. The latents of partially observed points are optimised with the training set, and then used to predict the missing values.
We assess the performance of the models using test-set perplexity (a measure of how much the model is surprised by the missing test values). This is defined as the exponent of the negative average log predicted probability.


\begin{figure}[b!]
\vspace{-4mm}
\subfloat[LGM]{
\hspace{-5mm}
  \includegraphics[trim = 0mm 0mm 0mm 0mm, clip, width=1.1\linewidth]{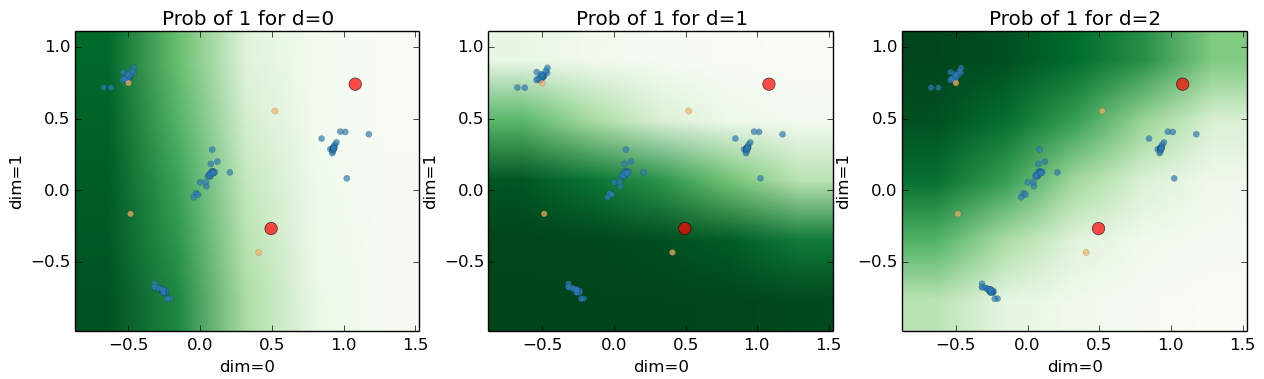}
} \vspace{-4mm}\\ 
\subfloat[CLGP]{
\hspace{-5mm}
  \includegraphics[trim = 0mm 0mm 0mm 0mm, clip, width=1.1\linewidth]{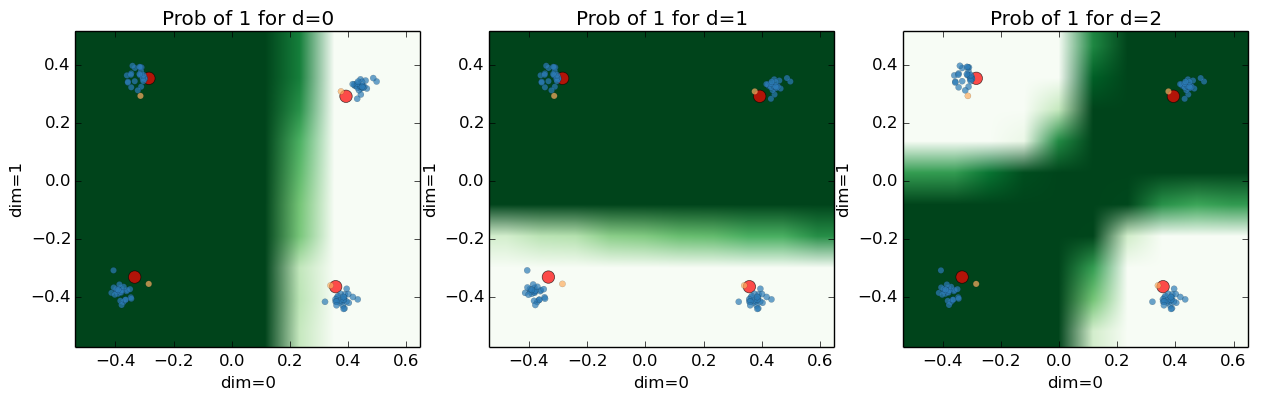}
}
\vspace{-3mm}
  \caption{\textbf{Density over the \textit{latent} space for XOR as predicted by the linear model (top, LGM), and non-linear model (bottom, CLGP).} Each figure shows the probability $p(\by_d = 1 | \bx)$ as a function of $\bx$, for $d=0,1,2$ (first, second, and third digits in the triplet left to right). The darker the shade of green, the higher the probability. In blue are the latents corresponding to the training points, in yellow are the latents corresponding to the four partially observed test points, and in red are the inducing points used to support the function.} \label{fig:XOR}
\end{figure}

\subsection{Linear Models Have Difficulty with Multi-modal Distributions}

We show the advantages of using a non-linear categorical distribution estimation compared to a linear one, evaluating the CLGP against the linear LGM. We implement the latent Gaussian model using a linear covariance function in our model; we remove the KL divergence term in $\bu$ following the model specification in \citep{khan2012stick}, and use our inference scheme described above. 
Empirically, the Monte Carlo inference scheme with the linear model results in the same test error on \citep{Takashi2008Asia} as the piece-wise bound based inference scheme developed in \citep{khan2012stick}. 

\subsubsection{Relation Embedding -- Exclusive Or}

A simple example of relational learning (following \citet{paccanaro2001learning}) can be used to demonstrate when linear latent space models fail. In this task we are given a dataset with example relations and the model is to capture the distribution that generated them. A non-linear dataset is constructed using the XOR (exclusive or) relation. We collect 25 positive examples of each assignment of the binary relation (triplets of the form $(0,0,0),~(0,1,1),~(1,0,1),~(1,1,0)$, corresponding to  $0~\text{XOR}~1 = 1$ and so on). We then maximise the variational lower bound using RMSPROP for both the linear and non-linear models with 20 samples for the Monte Carlo integration. 
We add four more triplets to the dataset: $(0,0,?),~(0,1,?),~(1,0,?),~(1,1,?)$. We evaluate the probabilities the models assign to each of the missing values (also known as \textit{imputation}) and report the results.

We assessed the test-set perplexity repeating the experiment 3 times and averaging the results. The linear model obtained a test-set perplexity (with standard deviation) of $75.785 \pm 13.221$, whereas the non-linear model obtained a test-set perplexity of $1.027 \pm 0.016$. A perplexity of 1 corresponds to always predicting correctly.

\begin{figure}[t]
\vspace{-6mm}
\hspace{-6mm}
\subfloat[LGM]{
\includegraphics[trim = 0mm 0mm 109mm 0mm, clip, width=0.35\linewidth]{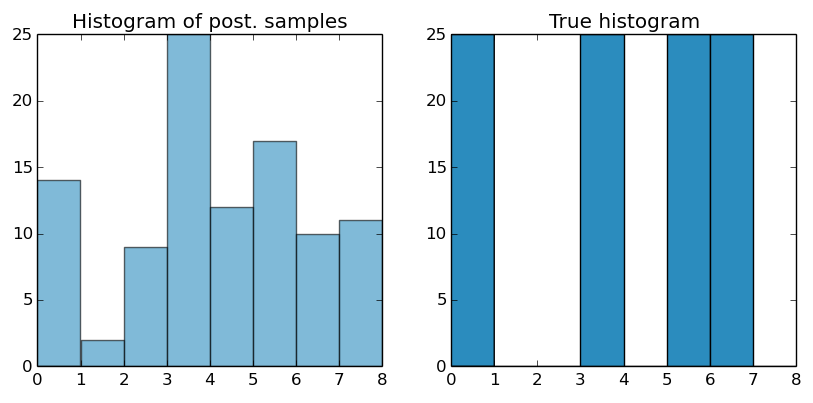}
}
\subfloat[CLGP]{
\includegraphics[trim = 0mm 0mm 105mm 0mm, clip, width=0.35\linewidth]{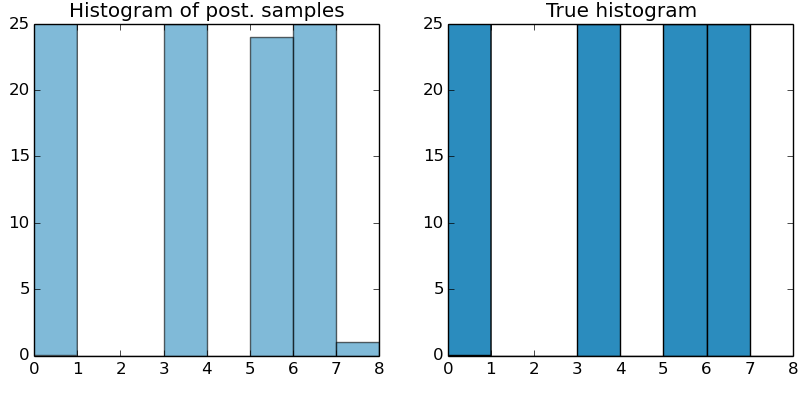}
}
\subfloat[Ground truth]{
\includegraphics[trim = 105mm 0mm 0mm 0mm, clip, width=0.35\linewidth]{figures/XOR_rbf_hist}
}
  \caption{\textbf{Histogram of categorical values for XOR} (encoded in binary for the 8 possible values) for samples drawn from the posterior of the latent space of the linear model (left, LGM), the non-linear model (middle, CLGP), and the data used for training (right).} \label{fig:XOR_hist}
  \vspace{-4mm}
\end{figure}

During optimisation the linear model jumps between different local modes, trying to capture all four possible triplets (fig.\ \ref{fig:XOR}). The linear model assigns probabilities to the missing values by capturing some of the triplets well, but cannot assign high probability to the others. 
In contrast, the CLGP model is able to capture all possible values of the relation. 
Sampling from probability vectors from the latent variational posterior for both models, we get a histogram of the posterior distribution (fig.\ \ref{fig:XOR_hist}). As can be seen, the CLGP model is able to fully capture the distribution whereas the linear model is incapable of it.

\subsubsection{Relation Learning -- Complex Example}

We repeat the previous experiment with a more complex relation. We generated 1000 strings from a simple probabilistic context free grammar with non-terminals $\{ \alpha, \beta, A, B, C \}$, start symbol $\alpha$, terminals $\{ a, b, c, d, e, f, g \}$, and derivation rules:
\begin{align*}
    \alpha &\rightarrow A \beta ~[1.0] \\
    \beta &\rightarrow B A ~[0.5] ~~|~~ C ~[0.5] \\
    A &\rightarrow a~ [0.5] ~~|~~ b ~[0.3] ~~|~~ c ~[0.2] \\
    B &\rightarrow d ~[0.7] ~~|~~ e ~[0.3] \\
    C &\rightarrow f ~[0.7] ~~|~~ g ~[0.3]
\end{align*}
where we give the probability of following a derivation rule inside square brackets. Following the start symbol and the derivation rules we generate strings of the form $ada$, $af$, $ceb$, and so on. We add two ``start'' and ``end'' symbols $s, \overline{s}$ to each string obtaining strings of the form $ssceb\overline{s}\overline{s}$. We then extract triplets of consecutive letters from these strings resulting in training examples of the form $ssc$, $sce$, ..., $b\overline{s}\overline{s}$. This is common practice in text preprocessing in the natural language community. It allows us to generate relations from scratch by conditioning on the prefix $ss$ to generate the first non-terminal (say $a$), then condition on the prefix $sa$, and continue in this vein iteratively. Note that a simple histogram based approach will do well on this dataset (and the previous dataset) as it is not sparse.

\begin{table}[h]
\vspace{-1mm}
\begin{center}
\begin{tabular}{|c|c|c|}
\hline 
Split & LGM & \textbf{CLGP} \\ 
\hline 
1 & $7.898 \pm 3.220$ & $2.769 \pm 0.070$ \\ 
\hline 
2 & $26.477 \pm 23.457$ & $2.958 \pm 0.265$ \\ 
\hline 
3 & $6.748 \pm 3.028$ & $2.797 \pm 0.081$ \\ 
\hline 
\end{tabular} 
\end{center}
\caption{\textbf{Test-set perplexity on relational data}. Compared are the linear LGM and the non-linear CLGP.}
\label{table:relation-embedding}
\vspace{-1mm}
\end{table}

We compared the test-set perplexity of the non-linear CLGP (with 50 inducing points) to that of the linear LGM on these training inputs, repeating the same experiment set-up as before. We impute one missing value at random in each test string, using $20\%$ of the strings as a test set with 3 different splits. The results are presented in table \ref{table:relation-embedding}. The linear model cannot capture this data well, and seems to get very confident about misclassified values (resulting in very high test-set perplexity). 

\subsection{Sparse Small Data}

\begin{table*}[t!]
\begin{center}
\begin{tabular}{|c|c|c|c|c|c|c|}
\hline 
Split & Baseline & Multinomial & Uni-Dir-Mult & Bi-Dir-Mult & LGM & \textbf{CLGP} \\ 
\hline 
1 & $8.68$ & $4.41$ & $4.41$ & $3.41$ & $3.57 \pm 0.208$ & $\mathbf{2.86 \pm 0.119}$ \\ 
\hline 
2 & $8.68$ & $\infty$ & $4.42$ & $\mathbf{3.49}$ & $\mathbf{3.47 \pm 0.252}$ & $\mathbf{3.36 \pm 0.186}$ \\ 
\hline 
3 & $8.85$ & $4.64$ & $4.64$ & $3.67$ & $12.13 \pm 9.705$ & $\mathbf{3.34 \pm 0.096}$ \\ 
\hline 
\end{tabular} 
\end{center}
\vspace{-2mm}
\caption{\textbf{Test-set perplexity on Breast cancer dataset, predicting randomly missing \textit{categorical test results}}. 
The models compared are: \textit{Baseline} predicting uniform probability for all values, \textit{Multinomial} -- predicting the probability for a missing value based on its frequency, \textit{Uni-Dir-Mult} -- Unigram Dirichlet Multinomial with concentration parameter $\alpha=0.01$, \textit{Bi-Dir-Mult} -- Bigram Dirichlet Multinomial with concentration parameter $\alpha=1$, \textit{LGM}, and the proposed model (\textit{CLGP}).}
\label{table:breast-cancer}
\vspace{-2mm}
\end{table*}

%

We assess the model in the real-world domain of small sparse data. We compare the CLGP model to a histogram based approach, demonstrating the difficulty with frequency models for sparse data. We further compare our model to the linear LGM.

\subsubsection{Medical Diagnosis}

We use the Wisconsin breast cancer dataset for which obtaining samples is a long and expensive process \citep{Zwitter1988Breast}. The dataset is composed of 683 data points, with 9 categorical variables taking values between 1 and 10, and an additional categorical variable taking 0,1 values -- $2 \times 10^9$ possible configurations. We use three quarters of the dataset for training and leave the rest for testing, averaging the test-set perplexity on three repetitions of the experiment.
We use three different random splits of the dataset as the distribution of the data can be fairly different among different splits. In the test set we randomly remove one of the 10 categorical values, and test the models' ability to recover that value. 
Note that this is a harder task than the usual use of this dataset for binary classification.
We use the same model set-up as in the first experiment. 

We compare our model (\textit{CLGP}) to a baseline predicting uniform probability for all
values (\textit{Baseline}), a multinomial model predicting the probability for a missing value based on its frequency (\textit{Multinomial}), a unigram Dirichlet Multinomial model with concentration parameter $\alpha = 0.01$ (\textit{Uni-DirMult}), a bigram Dirichlet Multinomial with concentration parameter $\alpha = 1$ (\textit{Bi-Dir-Mult}), and the linear continuous space model (\textit{LGM}).
More complicated frequency based approaches are possible, performing variable selection and then looking at frequencies of triplets of variables. These will be very difficult to work with in this sparse small data problem.

We evaluate the models' performance using the test-set perplexity metric as before (table \ref{table:breast-cancer}). As can be seen, the frequency based approaches obtain worse results than the continuous latent space approaches. The frequency model with no smoothing obtains perplexity $\infty$ for split 2 because one of the test points has a value not observed before in the training set. Using smoothing solves this. The baseline (predicting uniformly) obtains the highest perplexity on average. The linear model exhibits high variance for the last split, and in general has higher perplexity standard deviation than the non-linear model. 

\begin{figure*}[t!]
\hspace{-4mm}
\subfloat[Example alphadigits]{
  \includegraphics[trim = 12mm 12mm 10mm 10mm, clip, width=0.17\linewidth]{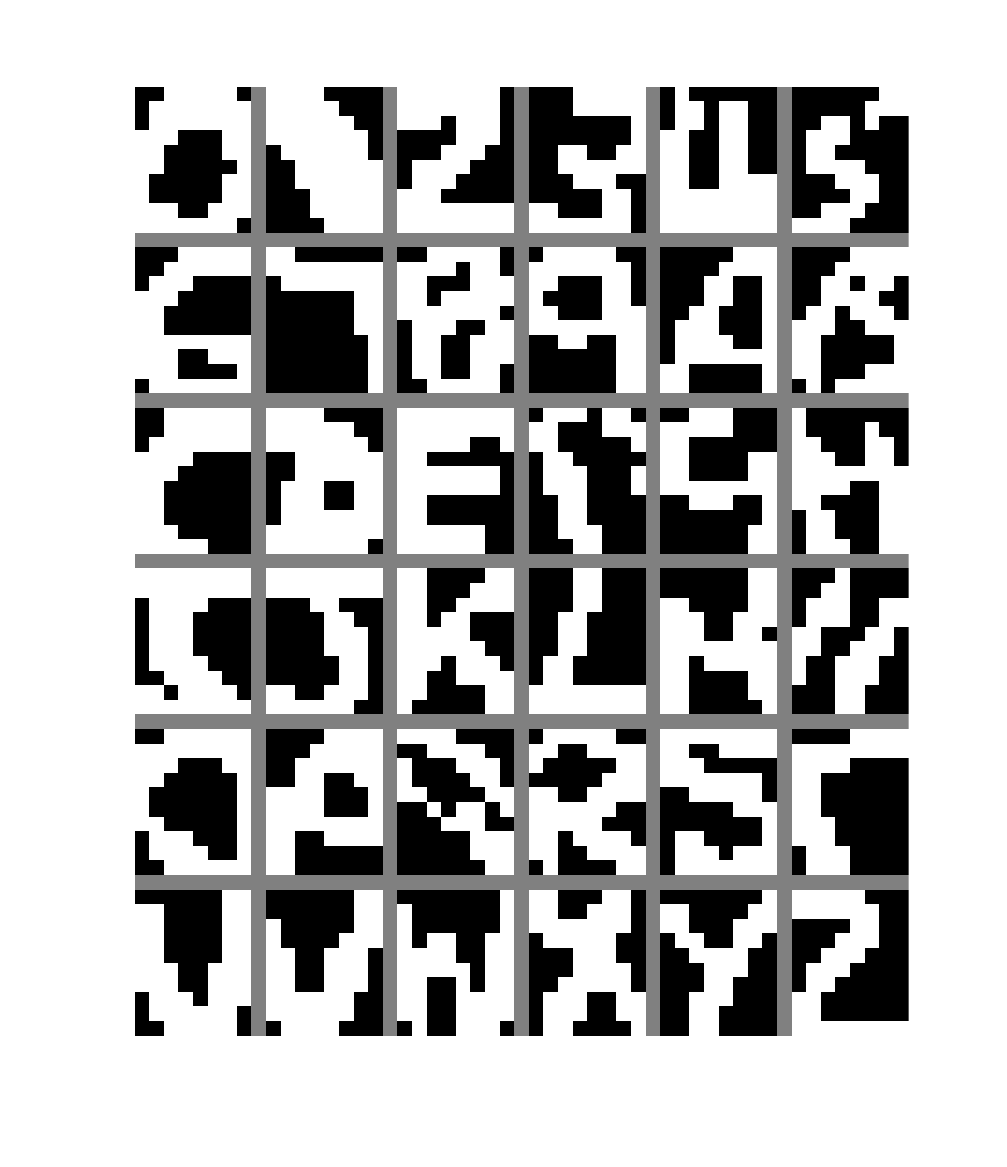}
  \label{fig:alphadigits_examples}
}
\subfloat[Train \& Test log perplexity]{
  \includegraphics[trim = 4mm 1mm 10mm 7mm, clip, width=0.28\linewidth]{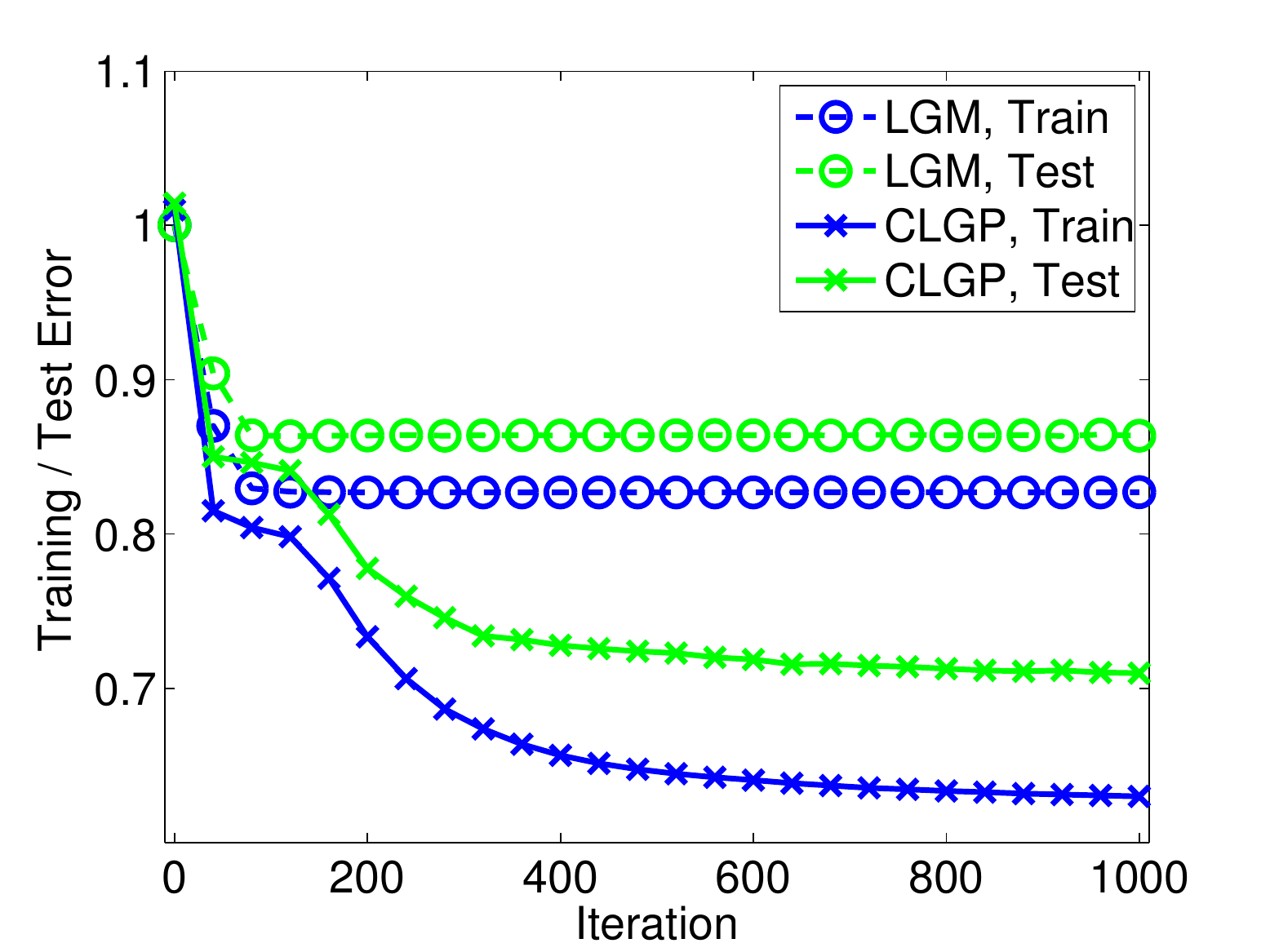}
  \label{fig:alphadigits_errors}
}
\subfloat[LGM latent space]{
  \includegraphics[trim = 1mm 1mm 14mm 10mm, clip, width=0.28\linewidth]{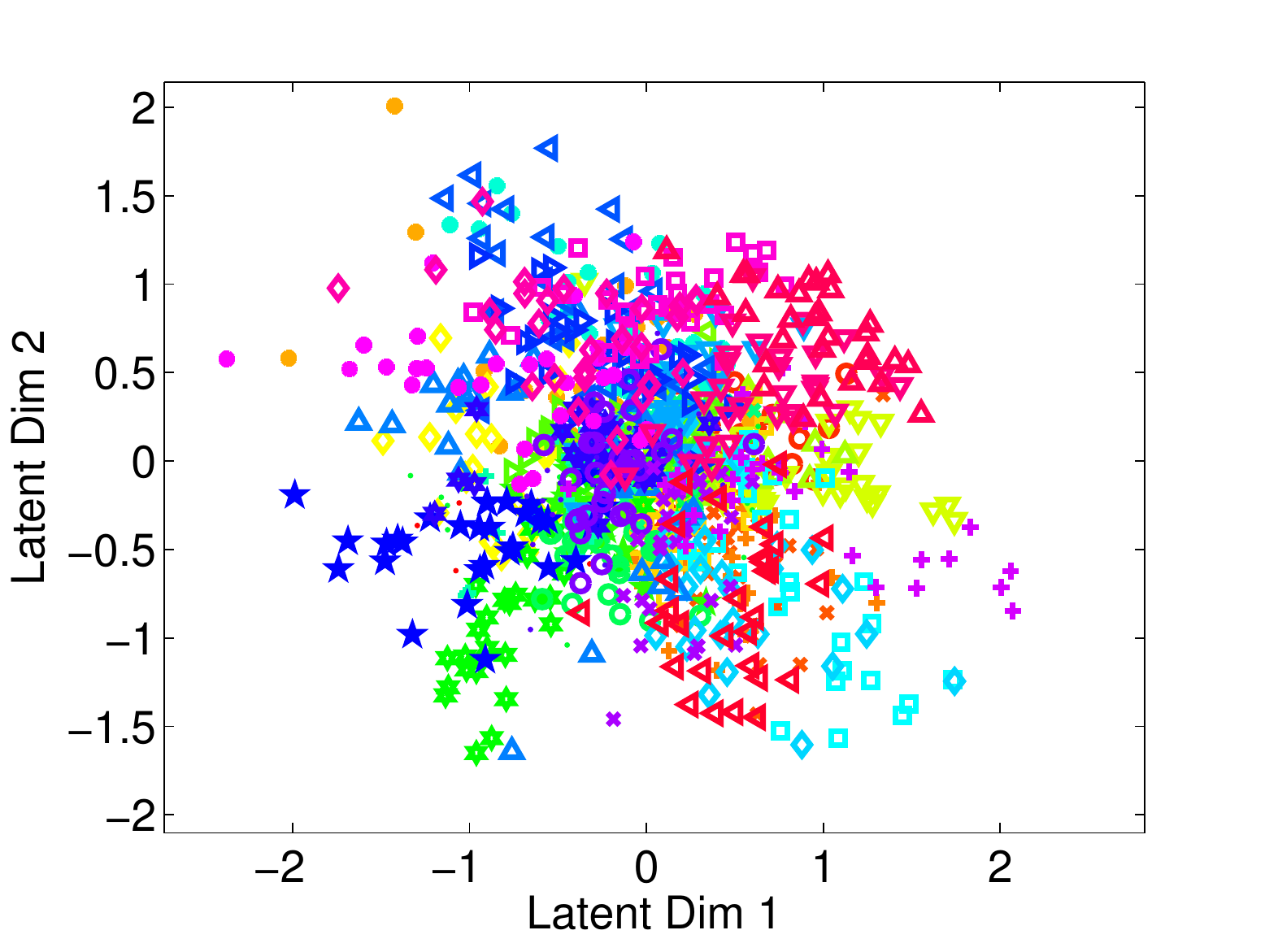}
  \label{fig:alphadigits_latents_LGM}
}
\subfloat[CLGP latent space]{
  \includegraphics[trim = 1mm 1mm 14mm 10mm, clip, width=0.28\linewidth]{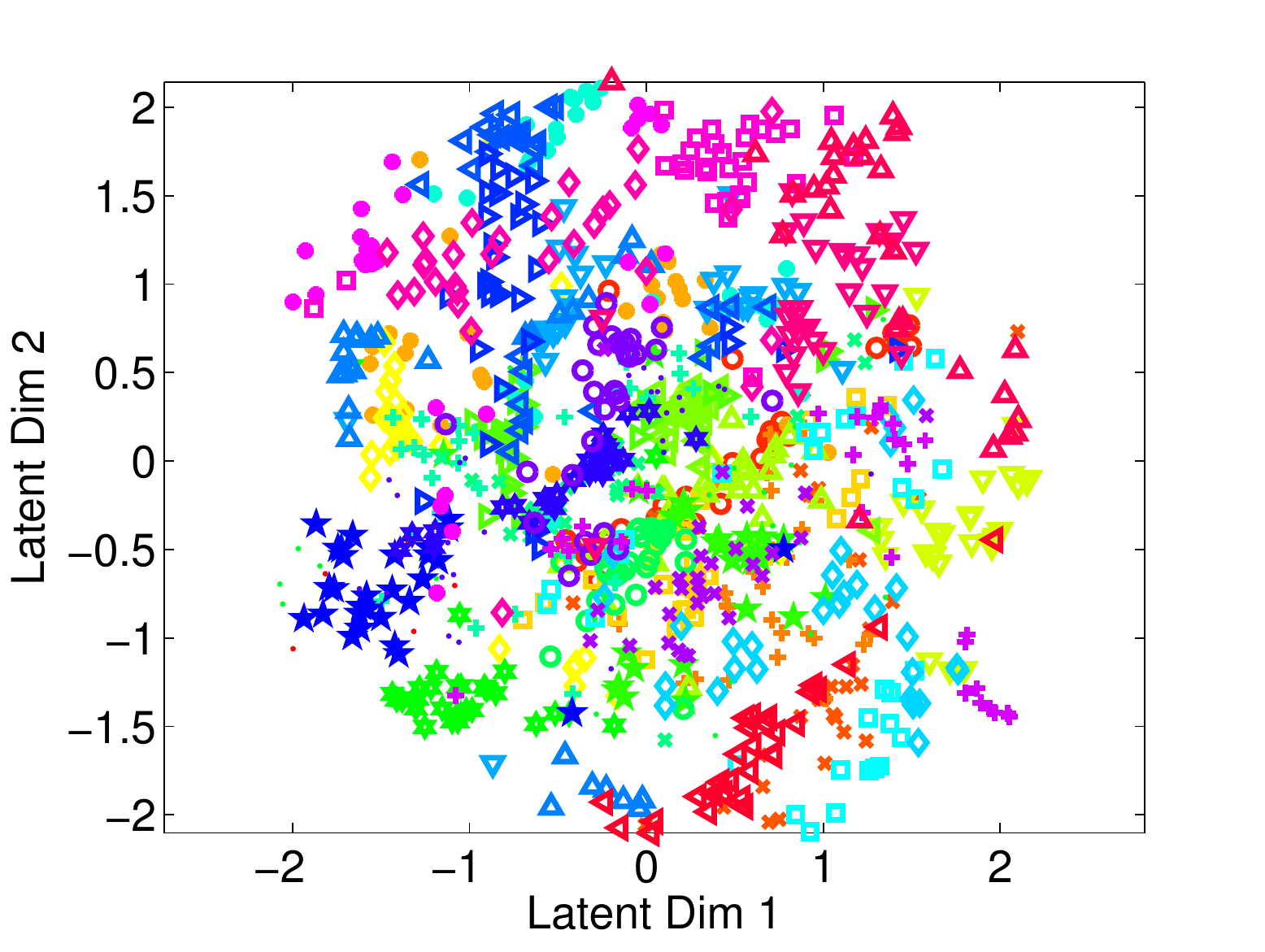}
  \label{fig:alphadigits_latents_CLGP}
}
  \caption{\textbf{Binary alphadigit dataset.} (a) example of each class (b) Train and test log perplexity for 1000 iterations. (c,d) 2-D latend embedding of the 36 alphadigit classes with LGM (left) and CLGP model (right). Each color-marker combination denotes one class.} \label{fig:alphadigits}
  \vspace{-2mm}
\end{figure*}

\subsubsection{Terror Warning Effects on Political Attitude}

We further compare the models on raw data collected in an experimental study of the effects of government terror warnings on political attitudes (from the START Terrorism Data Archive Dataverse\footnote{Obtained from the Harvard Dataverse Network \url{thedata.harvard.edu/dvn/dv/start/faces/study/StudyPage.xhtml?studyId=71190}}). 
This dataset consists of 1282 cases and 49 variables. 
We used 1000 cases for training and 282 for test.
From the 49 variables, we used the 17 corresponding to categorical answers to study questions (the other variables are subjects' geographic info and an answer to an open question). The 17 variables take 5 or 6 possible values for each variable with some of the values missing.

\begin{table}[h]
\vspace{-1mm}
\begin{center}
\begin{tabular}{|c|c|c|c|}
\hline 
Baseline & Multinomial & LGM & \textbf{CLGP} \\ 
\hline 
2.50 & 2.20 & $3.07$ & $\mathbf{2.11}$ \\ 
\hline 
\end{tabular} 
\end{center}
\caption{\textbf{Test-set perplexity for terror warning effects on political attitude}. Compared are the linear LGM and the non-linear CLGP.}
\label{table:Terror-Warning}
\vspace{-1mm}
\end{table}

We repeat the same experiment set-up as before, with a 6 dimensional latent dimensions, 100 inducing points, 5 samples to estimate the lower bound, and running the optimiser for 1000 iterations. We compare a \textit{Baseline} using a uniform distribution for all values, a \textit{Multinomial} model using the frequencies of the individual variables, the linear \textit{LGM}, and the \textit{CLGP}.
The results are given in table \ref{table:Terror-Warning}.
On the training set, the CLGP obtained a training error of $1.56$, and the LGM obtained a training error of $1.34$. The linear model seems to over-fit to the data. 

\subsubsection{Handwritten Binary Alphadigits}
We evaluate the performance of our model on a sparse dataset with a large number of dimensions. The dataset, Binary Alphadigits, is composed of $20 \times 16$ binary images of 10 handwritten digits and 26 handwritten capital letters, each class with 39 images\footnote{Obtained from \url{http://www.cs.nyu.edu/~roweis/data.html}}. 
We resize each image to $10 \times 8$, and obtain a dataset of 1404 data points each with 80 binary variables. 
We repeat the same experiment set-up as before with 2 latent dimensions for ease of visual comparison of the latent embeddings.
Fig.~\ref{fig:alphadigits_examples} shows an example of each alphadigit class. Each class is then randomly split to 30 training and 9 test examples. In the test set, we randomly remove $20\%$ of the pixels and evaluate the prediction error. 

Fig.~\ref{fig:alphadigits_errors} shows the training and test error in log perplexity for both models. LGM converges faster than CLGP but ends up with a much higher prediction error due to its limited modeling capacity with 2 dimensional latent variables. This is validated with the latent embedding shown in fig.~\ref{fig:alphadigits_latents_LGM} and \ref{fig:alphadigits_latents_CLGP}. Each color-marker combination denotes one alphadigit class. As can be seen, CLGP has a better separation of the different classes than LGM even though the class labels are not used in the training.

\begin{figure}[b!]
\vspace{-8mm}
   \center
\subfloat[LGM]{
  \includegraphics[trim = 3mm 3mm 2.5mm 2.5mm, clip, width=0.49\linewidth]{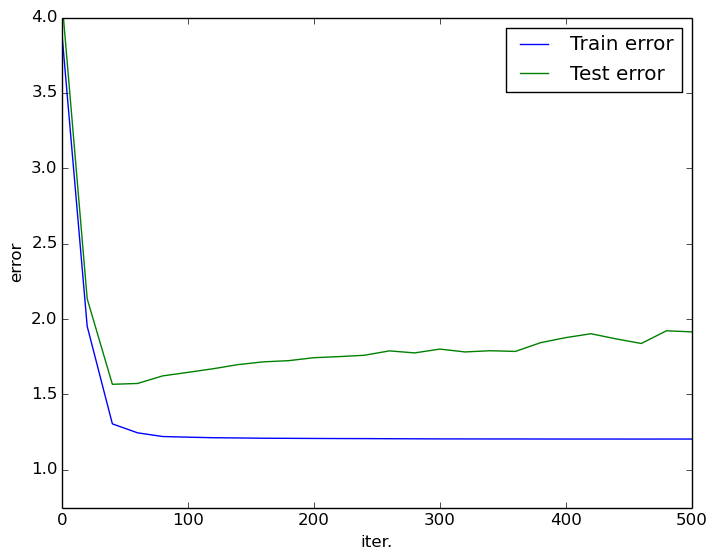}
}
\subfloat[CLGP]{
  \includegraphics[trim = 3mm 3mm 2.5mm 2.5mm, clip, width=0.49\linewidth]{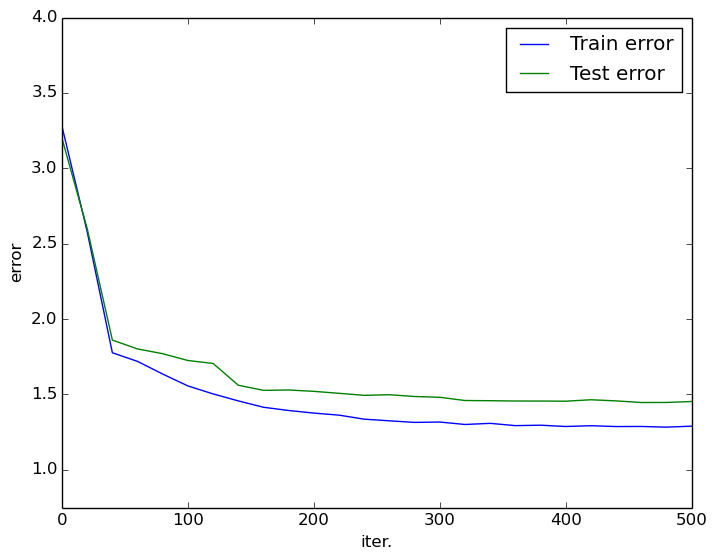}
}
  \caption{\textbf{Train and test log perplexities for LGM (left) and the CLGP model (right) for one of the splits of the breast cancer dataset.} The train log perplexity of LGM decreases while the test log perplexity starts increasing at iteration 50.} \label{fig:overfitting}
\end{figure}

\subsection{Latent Gaussian Model Over-fitting}

It is interesting to note that the latent Gaussian model (LGM) over-fits on the different datasets. It is possible to contribute this to the lack of regularisation over the linear transformation -- the weight matrix used to transform the latent space to the Softmax weights is optimised without a prior. In all medical diagnosis experiment repetitions, for example, the model's training log perplexity decreases while the test log perplexity starts increasing (see fig.\ \ref{fig:overfitting} for the train and test log perplexities of split 1 of the breast cancer dataset). Note that even though the test log perplexity starts increasing, at its lowest point it is still higher than the end test log perplexity of the CLGP model. This is observed for all splits and all repetitions.

It is worth noting that we compared the CLGP to the LGM on the ASES survey dataset \citep{Takashi2008Asia} used to assess the LGM in \citep{khan2012stick}. We obtained a test perplexity of 1.98, compared to LGM's test perplexity of 1.97. We concluded that the dataset was fairly linearly separable.

\subsection{Inference Robustness}

\begin{figure}[b!]
\hspace{-4mm}
  \includegraphics[trim = 7mm 0mm 14mm 6mm, clip,width=1.05\linewidth]{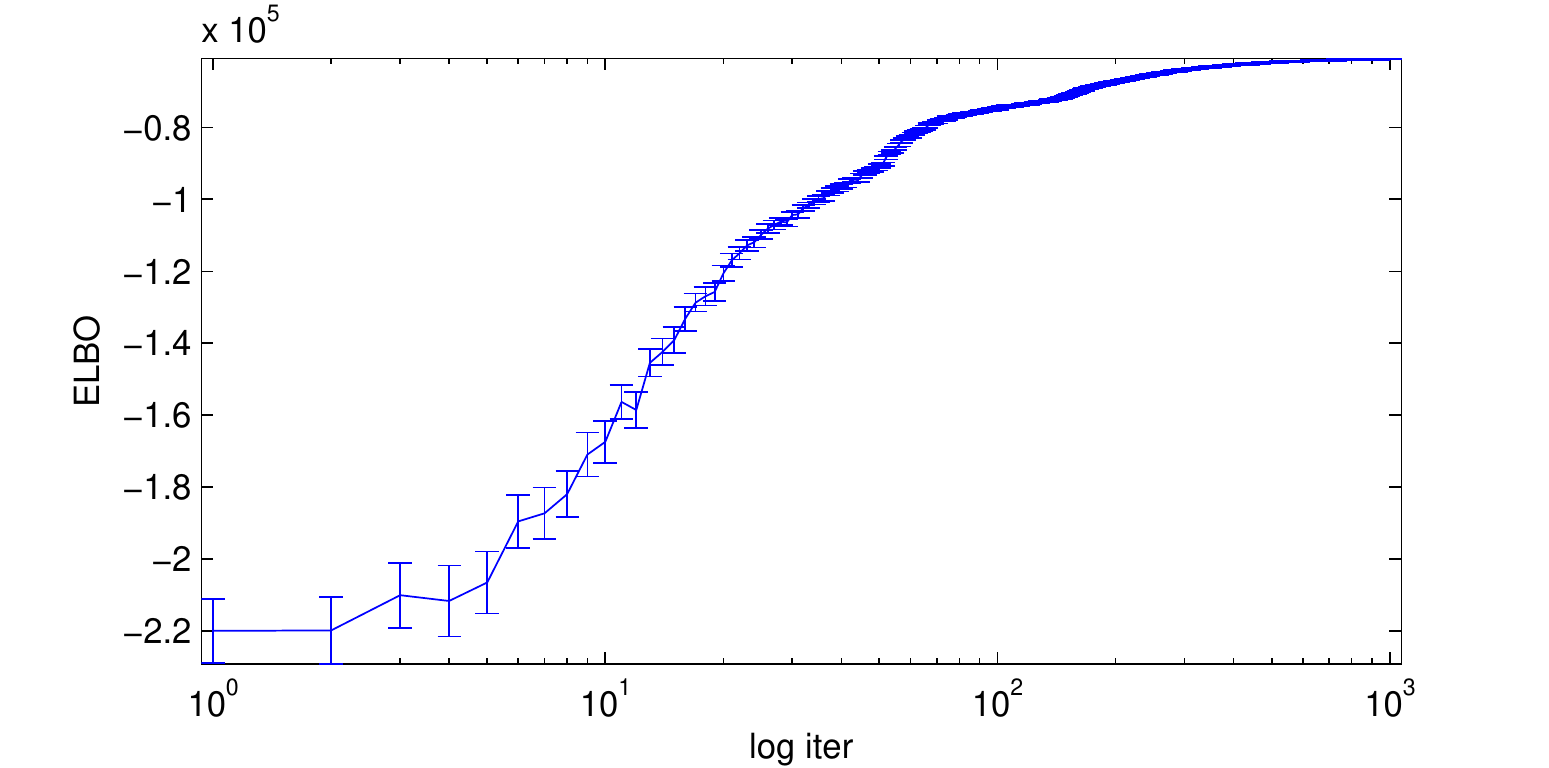}
  \caption{\textbf{ELBO and standard deviation per iteration} (on log scale) for the Alphadigits dataset.} \label{fig:standard_deviation}
\end{figure}

Lastly, we inspect the robustness of our inference, evaluating the Monte Carlo (MC) estimate standard deviation as optimisation progresses. Fig.\ \ref{fig:standard_deviation} shows the ELBO and MC standard deviation per iteration (on log scale) for the Alphadigit dataset. 
It seems that the estimator standard deviation decreases as the approximating variational distribution fits to the posterior. This makes the stochastic optimisation easier, speeding up convergence. A further theoretical study of this behaviour in future research will be of interest.

\section{Discussion and Conclusions}
We have presented the first Bayesian model capable of capturing sparse multi-modal categorical distributions based on a continuous representation. This model ties together many existing models in the field, linking the linear and discrete \textit{latent Gaussian models} to the non-linear continuous space \textit{Gaussian process latent variable model} and to the fully observed discrete \textit{Gaussian process classification}.


In future work we aim to answer short-comings in the current model such as scalability and robustness. We scale the model following research on GP scalability done in \citep{hensman2013gaussian,Gal2014DistributedB}. 
The robustness of the model depends on the sample variance in the Monte Carlo integration. As discussed in \citet{blei2012variational}, variance reduction techniques can help in the estimation of the integral, and methods such as the one developed in \citet{wang2013variance} can effectively increase inference robustness.


\bibliography{cgplvm}
\bibliographystyle{icml2015}

%% file: CLGP-appendix.tex
\appendix
\section{Appendix}
\subsection{Code}
The basic model and inference (without covariance matrix caching) can be implemented in 20 lines of Python and Theano for each categorical variable:

\begin{python}
import theano.tensor as T
m = T.dmatrix('m') # ..and other variables
X = m + s * randn(N, Q)
U = mu + L.dot(randn(M, K))
Kmm = RBF(sf2, l, Z)
Kmn = RBF(sf2, l, Z, X)
Knn = RBFnn(sf2, l, X)
KmmInv = T.matrix_inverse(Kmm)
A = KmmInv.dot(Kmn)
B = Knn - T.sum(Kmn * KmmInv.dot(Kmn), 0)
F = A.T.dot(U)+B[:,None]**0.5 * randn(N,K) 
S = T.nnet.softmax(F)
KL_U, KL_X = get_KL_U(), get_KL_X()
LS = T.sum(T.log(T.sum(Y * S, 1))) 
			- KL_U - KL_X 
LS_func = theano.function(['''inputs'''], 
			LS)
dLS_dm = theano.function(['''inputs'''], 
			T.grad(LS, m)) # and others
# ... and optimise LS with RMS-PROP
\end{python}

%% file: CLGP-arXiv.bbl
\begin{thebibliography}{29}
\providecommand{\natexlab}[1]{#1}
\providecommand{\url}[1]{\texttt{#1}}
\expandafter\ifx\csname urlstyle\endcsname\relax
  \providecommand{\doi}[1]{doi: #1}\else
  \providecommand{\doi}{doi: \begingroup \urlstyle{rm}\Url}\fi

\bibitem[Bengio et~al.(2006)Bengio, Schwenk, Sen{\'e}cal, Morin, and
  Gauvain]{bengio2006neural}
Bengio, Yoshua, Schwenk, Holger, Sen{\'e}cal, Jean-S{\'e}bastien, Morin,
  Fr{\'e}deric, and Gauvain, Jean-Luc.
\newblock Neural probabilistic language models.
\newblock In \emph{Innovations in Machine Learning}, pp.\  137--186. Springer,
  2006.

\bibitem[Bergstra et~al.(2010)Bergstra, Breuleux, Bastien, Lamblin, Pascanu,
  Desjardins, Turian, Warde-Farley, and Bengio]{bergstra+al:2010-scipy}
Bergstra, James, Breuleux, Olivier, Bastien, Fr{\'{e}}d{\'{e}}ric, Lamblin,
  Pascal, Pascanu, Razvan, Desjardins, Guillaume, Turian, Joseph, Warde-Farley,
  David, and Bengio, Yoshua.
\newblock Theano: a {CPU} and {GPU} math expression compiler.
\newblock In \emph{Proceedings of the Python for Scientific Computing
  Conference ({SciPy})}, June 2010.
\newblock Oral Presentation.

\bibitem[Blei \& Lafferty(2006)Blei and Lafferty]{blei2006correlated}
Blei, David and Lafferty, John.
\newblock Correlated topic models.
\newblock \emph{Advances in neural information processing systems},
  18:\penalty0 147, 2006.

\bibitem[Blei et~al.(2012)Blei, Jordan, and Paisley]{blei2012variational}
Blei, David~M, Jordan, Michael~I, and Paisley, John~W.
\newblock Variational {B}ayesian inference with stochastic search.
\newblock In \emph{Proceedings of the 29th International Conference on Machine
  Learning (ICML-12)}, pp.\  1367--1374, 2012.

\bibitem[B{\"o}hning(1992)]{bohning1992multinomial}
B{\"o}hning, Dankmar.
\newblock Multinomial logistic regression algorithm.
\newblock \emph{Annals of the Institute of Statistical Mathematics},
  44\penalty0 (1):\penalty0 197--200, 1992.

\bibitem[Collobert \& Weston(2008)Collobert and Weston]{collobert2008unified}
Collobert, Ronan and Weston, Jason.
\newblock A unified architecture for natural language processing: Deep neural
  networks with multitask learning.
\newblock In \emph{Proceedings of the 25th international conference on Machine
  learning}, pp.\  160--167. ACM, 2008.

\bibitem[Duchi et~al.(2011)Duchi, Hazan, and Singer]{duchi2011adaptive}
Duchi, John, Hazan, Elad, and Singer, Yoram.
\newblock Adaptive subgradient methods for online learning and stochastic
  optimization.
\newblock \emph{The Journal of Machine Learning Research}, 12:\penalty0
  2121--2159, 2011.

\bibitem[Gal et~al.(2014)Gal, van~der Wilk, and Rasmussen]{Gal2014DistributedB}
Gal, Yarin, van~der Wilk, Mark, and Rasmussen, Carl~E.
\newblock Distributed variational inference in sparse {G}aussian process
  regression and latent variable models.
\newblock In \emph{Advances in Neural Information Processing Systems 27}. 2014.

\bibitem[Hensman et~al.(2013)Hensman, Fusi, and Lawrence]{hensman2013gaussian}
Hensman, James, Fusi, Nicolo, and Lawrence, Neil~D.
\newblock Gaussian processes for big data.
\newblock \emph{arXiv preprint arXiv:1309.6835}, 2013.

\bibitem[Inoguchi(2008)]{Takashi2008Asia}
Inoguchi, Takashi.
\newblock Asia {E}urope survey ({ASES}): A multinational comparative study in
  18 countries, 2001.
\newblock In \emph{Inter-university Consortium for Political and Social
  Research (ICPSR)}, 2008.

\bibitem[Jaakkola \& Jordan(1997)Jaakkola and Jordan]{Jaakkola96avariational}
Jaakkola, Tommi~S. and Jordan, Michael~I.
\newblock A variational approach to {B}ayesian logistic regression models and
  their extensions.
\newblock In \emph{Proceedings of the Sixth International Workshop on
  Artificial Intelligence and Statistics}, 1997.

\bibitem[Khan et~al.(2012)Khan, Mohamed, Marlin, and Murphy]{khan2012stick}
Khan, Mohammad~E, Mohamed, Shakir, Marlin, Benjamin~M, and Murphy, Kevin~P.
\newblock A stick-breaking likelihood for categorical data analysis with latent
  {G}aussian models.
\newblock In \emph{International conference on Artificial Intelligence and
  Statistics}, pp.\  610--618, 2012.

\bibitem[Kingma \& Welling(2013)Kingma and Welling]{kingma2013auto}
Kingma, Diederik~P and Welling, Max.
\newblock Auto-encoding variational {B}ayes.
\newblock \emph{arXiv preprint arXiv:1312.6114}, 2013.

\bibitem[Knowles \& Minka(2011)Knowles and Minka]{knowles2011non}
Knowles, David~A and Minka, Tom.
\newblock Non-conjugate variational message passing for multinomial and binary
  regression.
\newblock In \emph{Advances in Neural Information Processing Systems}, pp.\
  1701--1709, 2011.

\bibitem[Lawrence(2005)]{lawrence2005probabilistic}
Lawrence, Neil.
\newblock Probabilistic non-linear principal component analysis with {G}aussian
  process latent variable models.
\newblock \emph{The Journal of Machine Learning Research}, 6:\penalty0
  1783--1816, 2005.

\bibitem[Marlin et~al.(2011)Marlin, Khan, and Murphy]{marlin2011piecewise}
Marlin, Benjamin~M, Khan, Mohammad~Emtiyaz, and Murphy, Kevin~P.
\newblock Piecewise bounds for estimating bernoulli-logistic latent {G}aussian
  models.
\newblock In \emph{ICML}, pp.\  633--640, 2011.

\bibitem[Paccanaro \& Hinton(2001)Paccanaro and Hinton]{paccanaro2001learning}
Paccanaro, Alberto and Hinton, Geoffrey~E.
\newblock Learning distributed representations of concepts using linear
  relational embedding.
\newblock \emph{Knowledge and Data Engineering, IEEE Transactions on},
  13\penalty0 (2):\penalty0 232--244, 2001.

\bibitem[Qui{\~n}onero-Candela \& Rasmussen(2005)Qui{\~n}onero-Candela and
  Rasmussen]{quinonero2005unifying}
Qui{\~n}onero-Candela, Joaquin and Rasmussen, Carl~Edward.
\newblock A unifying view of sparse approximate {G}aussian process regression.
\newblock \emph{The Journal of Machine Learning Research}, 6:\penalty0
  1939--1959, 2005.

\bibitem[Rezende et~al.(2014)Rezende, Mohamed, and
  Wierstra]{rezende2014stochastic}
Rezende, Danilo~J, Mohamed, Shakir, and Wierstra, Daan.
\newblock Stochastic backpropagation and approximate inference in deep
  generative models.
\newblock In \emph{Proceedings of the 31st International Conference on Machine
  Learning (ICML-14)}, pp.\  1278--1286, 2014.

\bibitem[Schaul et~al.(2013)Schaul, Antonoglou, and Silver]{SchaulAS13}
Schaul, Tom, Antonoglou, Ioannis, and Silver, David.
\newblock Unit tests for stochastic optimization.
\newblock abs/1312.6055, 2013.
\newblock URL \url{http://arxiv.org/abs/1312.6055}.

\bibitem[Tieleman \& Hinton(2012)Tieleman and Hinton]{Tieleman2012COURSERA}
Tieleman, T. and Hinton, G.
\newblock Lecture 6.5 - rmsprop, {COURSERA}: Neural networks for machine
  learning, 2012.

\bibitem[Tipping \& Bishop(1999)Tipping and Bishop]{tipping1999probabilistic}
Tipping, Michael~E and Bishop, Christopher~M.
\newblock Probabilistic principal component analysis.
\newblock \emph{Journal of the Royal Statistical Society: Series B (Statistical
  Methodology)}, 61\penalty0 (3):\penalty0 611--622, 1999.

\bibitem[Titsias \& Lawrence(2010)Titsias and Lawrence]{titsias2010bayesian}
Titsias, Michalis and Lawrence, Neil.
\newblock Bayesian {G}aussian process latent variable model.
\newblock In \emph{Thirteenth International Conference on Artificial
  Intelligence and Statistics (AISTATS)}, 2010.

\bibitem[Titsias \& L{\'a}zaro-Gredilla(2014)Titsias and
  L{\'a}zaro-Gredilla]{titsias2014doubly}
Titsias, Michalis and L{\'a}zaro-Gredilla, Miguel.
\newblock Doubly stochastic variational {B}ayes for non-conjugate inference.
\newblock In \emph{Proceedings of the 31st International Conference on Machine
  Learning (ICML-14)}, pp.\  1971--1979, 2014.

\bibitem[Titsias(2009)]{Titsias09variationallearning}
Titsias, Michalis~K.
\newblock Variational learning of inducing variables in sparse {G}aussian
  processes.
\newblock In \emph{Artificial Intelligence and Statistics 12}, pp.\  567--574,
  2009.

\bibitem[Wang et~al.(2013)Wang, Chen, Smola, and Xing]{wang2013variance}
Wang, Chong, Chen, Xi, Smola, Alex, and Xing, Eric.
\newblock Variance reduction for stochastic gradient optimization.
\newblock In \emph{Advances in Neural Information Processing Systems}, pp.\
  181--189, 2013.

\bibitem[Williams \& Rasmussen(2006)Williams and
  Rasmussen]{rasmussen2006gaussian}
Williams, Christopher~KI and Rasmussen, Carl~Edward.
\newblock Gaussian processes for machine learning.
\newblock \emph{the MIT Press}, 2\penalty0 (3):\penalty0 4, 2006.

\bibitem[Zeiler(2012)]{zeiler2012adadelta}
Zeiler, Matthew~D.
\newblock Adadelta: An adaptive learning rate method.
\newblock \emph{arXiv preprint arXiv:1212.5701}, 2012.

\bibitem[Zwitter \& Soklic(1988)Zwitter and Soklic]{Zwitter1988Breast}
Zwitter, M. and Soklic, M.
\newblock Breast cancer dataset.
\newblock In \emph{University Medical Centre, Institute of Oncology, Ljubljana,
  Yugoslavia}, 1988.

\end{thebibliography}
